\documentclass{article}

\usepackage[preprint]{neurips_2020}

\usepackage{xcolor}
\usepackage{colortbl}
\definecolor{linkcolor}{RGB}{74, 102, 146}

\usepackage[utf8]{inputenc}
\usepackage[T1]{fontenc}
\usepackage[colorlinks=true,allcolors=linkcolor,pageanchor=true,plainpages=false,pdfpagelabels,bookmarks,bookmarksnumbered]{hyperref}

\usepackage{url}            
\usepackage{booktabs}       
\usepackage{amsfonts}       
\usepackage{nicefrac}       
\usepackage{microtype}      
\usepackage{nicefrac}

\DeclareSymbolFont{bbold}{U}{bbold}{m}{n}
\DeclareSymbolFontAlphabet{\mathbbold}{bbold}

\usepackage[english]{babel}

\usepackage{graphicx}
\usepackage{animate}
\usepackage{epsfig} 				
\usepackage{epstopdf}

\usepackage{listings}
\usepackage{color}
\usepackage{nameref}
\usepackage{hyperref}
\usepackage{amsmath}	 			
\usepackage{amssymb}  				

\usepackage{dsfont}			
\usepackage{mathtools}

\usepackage{epigraph}
\usepackage{lscape}
\usepackage[]{nomencl}				
\usepackage{algorithm}
\usepackage{algorithmic}
\usepackage{multicol}
\usepackage{multirow}
\usepackage{etoolbox}

\usepackage{caption}
\usepackage{subcaption}
\usepackage{wrapfig}

\usepackage{siunitx}

\usepackage{floatflt}

\usepackage{url}

\usepackage{enumitem}

\usepackage[frozencache,cachedir=minted-cache]{minted}
\usepackage{inconsolata}
\usepackage[T1]{fontenc}
\usepackage{xspace}
\newcommand{\eg}{{\it e.g.}\xspace}
\newcommand{\ie}{{\it i.e.}\xspace}

\newcommand{\email}[1]{\href{mailto:#1}{\nolinkurl{#1}}}

\newcommand{\fig}[1]{Fig.~\ref{#1}}
\newcommand{\eq}[1]{Equation~\eqref{#1}}

\newcommand{\namelib}[0]{\protect{MBRL-Lib}}
\newcommand{\titlelong}[0]{\namelib{}: A Modular Library for \\Model-based Reinforcement Learning}
\newcommand{\website}[0]{\url{https://github.com/facebookresearch/mbrl-lib}}

\newcommand{\onedmodel}[0]{\lstinline{OneDimTransitionRewardModel}}
\newcommand{\lst}[1]{\lstinline{#1}}

\newcommand{\mdptime}[0]{t}
\newcommand{\mdpstate}[0]{s}
\newcommand{\mdpaction}[0]{a}

\newcommand{\state}[0]{\mdpstate_\mdptime}
\newcommand{\action}[0]{a_t}
\newcommand{\hor}[0]{h}
\newcommand{\dynmod}[0]{f}

\newcommand{\E}[0]{\mathbb{E}} 				
\newcommand{\D}[0]{\mathcal{D}} 				
\newcommand{\sR}{\mathbb{R}}
\newcommand{\gI}{\mathcal{I}}

\newcommand{\MDPdimState}[0]{n}
\newcommand{\MDPdimAction}[0]{m}

\DeclareMathOperator*{\argmin}{arg\,min}
\DeclareMathOperator*{\argmax}{arg\,max}

\newcommand{\defeq}{\vcentcolon=}

\title{\titlelong}

\author{%
  Luis Pineda \and Brandon Amos \and Amy Zhang \and Nathan O. Lambert \and Roberto Calandra  \\
  Facebook AI Research\\
  University of California, Berkeley\\
  \texttt{\{lep,bda,amyzhang,rcalandra\}@fb.com, nol@berkeley.edu} \\
}

\definecolor{bdaemph}{RGB}{168, 141, 201}

\begin{document}

\maketitle


\begin{abstract}
	Model-based reinforcement learning is a compelling framework for data-efficient learning of agents that interact with the world.
This family of algorithms has many subcomponents that need to be carefully selected and tuned.
As a result the entry-bar for researchers to approach the field and to deploy it in real-world tasks can be daunting.
In this paper, we present \namelib{} -- a machine learning library for model-based reinforcement learning in continuous state-action spaces based on PyTorch.
\namelib{} is designed as a platform for both researchers, to easily develop, debug and compare new algorithms, and non-expert user, to lower the entry-bar of deploying state-of-the-art algorithms.
\namelib{} is open-source at \website{}.


\end{abstract}


\section{Introduction}

	Model-based Reinforcement Learning (MBRL) for continuous control is a
growing area of research
that investigates agents explicitly modeling and interacting with the world.
MBRL is capable of learning to control rapidly from a
limited number of trials, and enables us to integrate specialized domain
knowledge into the agent about how the world works.
While methods are largely exploratory and unsettled, some shared
components of these systems are emerging that work in conjunction
and harmony together:
the forward dynamics model, the
state-action belief propagation, the reward function, and the
planner/policy optimizer.

Albeit promising, MBRL suffers from reproducibility and
computational issues stemming from the broader issues
in the field \citep{henderson2017rlmatters}.
Moreover, MBRL methods can be even more difficult to debug and get working due to the
depth of individual components and complex interactions between
them, \eg{} between the model losses and control objectives
\citep{lambert2020mismatch}.

In this work we present the library \namelib{} to provide shared foundations, tools,
abstractions, and evaluations for continuous-action MBRL, providing:
1) software abstractions for lightweight and modular MBRL
for practitioners and researchers,
2) debugging and visualization tools
3) re-implementations of the state-of-the-art MBRL methods
that can be easily modified and extended.
Our goal is to see \namelib{} grow as an open source project where new
algorithms will be developed with this library and then added back.
Our hope is that with these established baselines, growth in MBRL
research will match the recent progress seen in model-free methods.



\section{Related Work on Reinforcement Learning Software}
\label{sec:related}

Many reinforcement learning libraries have been publicly released~\citep{Dhariwal2017OpenAI,pytorchrl,TFAgents} that provide high-quality implementation of the complex components involved in training RL agents. 
However, the vast majority of this focus on model-free reinforcement learning and lack crucial components of implementing MBRL algorithms.
The availability of these libraries has opened up model-free RL methods to researchers and practitioners alike, 
providing opportunities to researchers that we believe contribute to the noted empirical gap in performance between model-free and model-based methods -- in spite of theoretical evidence that model-based methods are superior in sample complexity~\citep{tu2018gap}. 

In MBRL, the code landscape consists mostly of a relatively limited number of specific algorithm implementations that are publicly available~\citep{Chua2018Deep,janner2019mbpo,wang2019exploring}.
Yet, none of these implementations provide a general and flexible environment that span multiple algorithms.
A closely related effort is \citep{wang2019benchmarking}, which provides a unified interface for benchmarking MBRL methods, but it does not provide a unified code base and it is not designed to facilitate future research.

To foster the development of novel and impactful methods, substantial development and software design are required. 
Aside from \namelib{}, there are two other notable efforts to create centralized libraries for MBRL: Baconian~\citep{linsen2019baconian} and Bellman~\citep{bellman2021}, both of which are implemented in Tensorflow~\citep{abadi2016tensorflow}.
Bellman, in particular, has been developed concurrently with \namelib{} and includes implementations for recent algorithms such as PETS, MBPO, and ME-TRPO, some of which are also included in \namelib{}.
In contrast to these libraries, \namelib{} is the first MBRL
library built to facilitate research in PyTorch~\citep{pytorch2019}.



\section{Background on Continuous Model-based Reinforcement Learning}
\label{sec:background}

	Here we briefly discuss the foundations of model-based reinforcement learning
for \emph{continuous control} of \emph{single-agent systems}
and cover problem formulation, modeling the transition dynamics,
and planning or policy optimization.

\paragraph{Problem Formulation}
MBRL often formulates the prediction problem as a Markov Decision Process (MDP)~\citep{bellman1957markovian}.
We consider MDPs with continuous states $\state \in \mathbb{R}^\MDPdimState$, continuous actions  $\action \in \mathbb{R}^{\MDPdimAction}$, a reward function $r(\state,\action)\in \mathbb{R}$, and a transition function $p : \mathbb{R}^{\MDPdimState \times \MDPdimAction \times \MDPdimState} \mapsto [0,\infty)$ representing the probability density of transitioning to state $\mdpstate_t$ given that action $\mdpaction$ was selected at state $\mdpstate_t$.\footnote{Problems with high-dimensional observations, both continuous and discrete, \eg pixels, are within the target scope of \namelib{}. Here we focus on the continuous-state case to simplify the presentation.}
A solution to a finite-time MDP is a policy/controller, $\pi^*(\cdot)$, mapping environment states to distributions over actions, which maximizes the expected sum of rewards:
\begin{equation}
    \pi^* = \argmax_{\pi}\mathbb{E}_{s_{t+1} \sim p(s_t,a_t), a_t \sim \pi(s_t)} \Big[\sum_{t=0}^{T} r(s_t, a_t) \Big]\,.
\end{equation}

\paragraph{Modeling and Learning the Transition Dynamics}
MBRL agents learn a forward dynamics model 
\begin{equation}
    \hat{\mdpstate}_{t+1} \sim \dynmod_\theta(\mdpstate_t, \mdpaction_t)\,,
    \label{eq:forward_mod}
\end{equation}
to approximate the true transition function of the environment; often, rewards function and termination functions are also learned, which we omit here for clarity.
Frequently, when learning the transition function, one can also model the delta transition between the current and next states as
\begin{equation}
    \hat{\mdpstate}_{t+1} \sim \mdpstate_t + \dynmod_\theta(\mdpstate_t, \mdpaction_t)\,.
    \label{eq:delta_obs}
\end{equation}

The model is trained on a dataset~$\D=\{(\mdpstate_i, \mdpaction_i, \mdpstate_{i+1}, r_i, d_i) \}_{i=1}^N$, where $r_i$ is a sampled reward and $d_i$ is a termination indicator denoting the end of the episode.
How to best collect this dataset is an open research question.
In practice, it is often collected by a mix of random exploration
followed by iterative control policies that optimize within simulated
model dynamics.

An important modeling choice is whether the learned model is
deterministic or probabilistic over the next state predictions.
For \emph{deterministic models}, the standard approach is to train the model
to minimize Mean Square Error (MSE) between the predicted and true
states, as
\begin{align}
l_\text{MSE} &= \sum_{n=1}^N \| \dynmod_\theta(\mdpstate_n, \mdpaction_n) - \mdpstate_{n+1} \|^2_2 \,.
\label{eq:mse}
\end{align}

\emph{Probabilistic models}
typically optimize the Negative Log Likelihood (NLL).
For example, a Gaussian policy predicts a distribution over
next states as
$\hat{\mdpstate}_{t+1} \sim \mathcal{N}\big(\mu_\theta(\mdpstate_t, \mdpaction_t), \Sigma_\theta(\mdpstate_t, \mdpaction_t) \big))$,
and the NLL is
\begin{align}
l_\text{NLL} & =  \sum_{n=1}^N  [\mu_\theta (\mdpstate_n, \mdpaction_n) - \mdpstate_{n+1}]^T \Sigma_{\theta}^{-1} (\mdpstate_n, \mdpaction_n) [\mu_\theta (\mdpstate_n, \mdpaction_n) - \mdpstate_{n+1}] + \text{log det } \Sigma_\theta (\mdpstate_n, \mdpaction_n).
\label{eq:nll}
\end{align}
\citet{Chua2018Deep, janner2019mbpo} incorporate uncertainty over model predictions, \ie \emph{epistemic} uncertainty, by using ensembles of independently trained models rather than a single model.
These ensembles are typically trained with
bootstrapping~\citep{efron1994introduction} and involves
training each model $\dynmod_{\theta_j}$ on its own copy of the dataset, $\mathcal{D}_j$, sampled with replacement from the original dataset $\mathcal{D}$; all models are trained independently.
The final prediction for an ensemble of $M$ models is
\begin{align}
    \hat{\mdpstate}_{t+1} = \frac{1}{M} \sum_{j=1}^M\dynmod_{\theta_j}(\mdpstate_t, \mdpaction_t) .
    \label{eq:ensemble_mean}
\end{align}

\paragraph{Planning or Policy Optimization}
Once a forward dynamics model is available, the next step is to use some planning or policy optimization algorithm to compute the next action as
\begin{align}
    a_{t+1} \sim \pi(s_t, \dynmod)\,.
\end{align}
In MBRL, policies are obtained by maximizing the sum of
expected rewards over a predictive horizon.
Due to approximation errors, long horizon prediction is not yet
practical in MBRL, so, many methods choose the next action to take by
unrolling the model predictions over a finite horizon $\hor$, which
can be much shorter than the MDP horizon $T$.
Here, the goal for a MBRL controller is to maximize the
predicted expected return over the horizon. For deterministic policies, this can be done by solving
\begin{align}
    a_{t:t+h}^* & = \argmax_{a_{t:t+h}}\; \sum_{i=0}^{\hor-1} \mathbb{E}_{\hat s_{t+i}} \left[ r(\hat{\mdpstate}_{t+i}, a_{t+i}) \right]\,,
    \label{eq:control}
\end{align}
wherein the expectation over the future states $\hat s_t$
is usually computed numerically by sampling some number of trajectories from the model.
In the case of ensembles, rather than using the mean ensemble prediction as in Eq.~(\ref{eq:ensemble_mean}), it is useful to propagate the uncertainty by uniformly sampling a model from the ensemble, and then sample a prediction from the sampled member of the ensemble.

Optimizing Eq.~(\ref{eq:control}) is a key distinction between MBRL methods.
Some popular approaches are:

\begin{itemize}[leftmargin=*,noitemsep,topsep=0pt]
    \item \textbf{Trajectory sampling and controller distillation:} A set of open-loop trajectories are sampled independently, and the best one is chosen~\citep{Chua2018Deep,hafner2019learning,lambert2019low,nagabandi2020deep}.
      Popular methods are random sampling and the Cross-Entropy Method
      (CEM)~\citep{de2005tutorial} and extensions as in
      \citet{wang2019exploring,amos2020dcem,nagabandi2020deep}.
    Often, to avoid overfitting to model idiosyncrasies, it's
beneficial to choose the mean over a set of top trajectories, rather
than the best trajectory itself.
Furthermore,
\citet{levine2013guided,weber2017imagination,pascanu2017learning,byravan2019imagined,lowrey2018plan} guide or condition policy training with
model-based rollouts.
    \item \textbf{Differentiate through the model:} For non ensemble models, one can use a parameterized policy together with automatic differentiation through the dynamics model to find actions that maximize model performance~\citep{deisenroth2013gaussian,levine2013guided,heess2015learning,henaff2018model,byravan2019imagined,amos2021svg}.
    \item \textbf{Use a model-free policy: } In this case a model-free learner, such as SAC~\citep{haarnoja2018soft}, is used over the predicted dynamics, often by populating a populating a replay buffer with ``imagined'' trajectories obtained from the model~\citep{gu2016continuous,kurutach2018model,janner2019mbpo}.
\end{itemize}

Some of these methods can also be used in a Model Predictive Control
(MPC) fashion~\citep{camacho2013model}, wherein a full sequence of
actions is obtained at each step, but only the first action in the
computed trajectory is chosen; this is the standard approach when
using trajectory sampling-based methods.
The above is also not a comprehensive list of the wide variety of
existing methods for generating continuous-action policies over a learned model.

\paragraph{Model-based methods in discrete systems}
We have focused this section on continuous spaces and
refer to \citet{kaiser2019model,schrittwieser2020mastering,hamrick2020role}
for some starter references on model-based methods
in discrete spaces.



\section{Software Architecture}  
\label{sec:approach}

	In this section we detail the various design choices made and relevant modules needed to implement different MBRL algorithms.

\subsection{Design Choices}

We designed \namelib{} with few well-defined principles of MBRL development in mind:
\begin{itemize}[leftmargin=*,noitemsep,topsep=0pt]
    \item \textbf{Modularity} \, MBRL algorithms involve the interplay of multiple components whose internal operations are often hidden from each other.
    Therefore, once such an algorithm has been written, it should be possible to alter some of the component choices (\eg, replace the trajectory sampling method) without affecting the others.
    Thus, \namelib{} seeks to enable a ``mix-and-match'' approach, where new algorithms or variants of existing ones can be easily written and tested without a lot of code involved.
    \item \textbf{Ease-of-use} \, We favor minimizing the amount of code required for users of the library, by heavy use of configuration files and a large set of utilities to encapsulate common operations, with an effort to make them general enough to accommodate new functionality. All functions are thoroughly documented and we provide examples of use. Using the library should be as friction-less as possible.
    \item \textbf{Performance} \, We consider crucial to provide high performance in terms of sample complexity and running time, and well-tuned hyperparameters for provided algorithms. More importantly, when performance gaps exists, future improvements to our components should be transparent to the end-users.
\end{itemize}

\begin{figure}
    \centering
    \includegraphics*[width=0.8\textwidth]{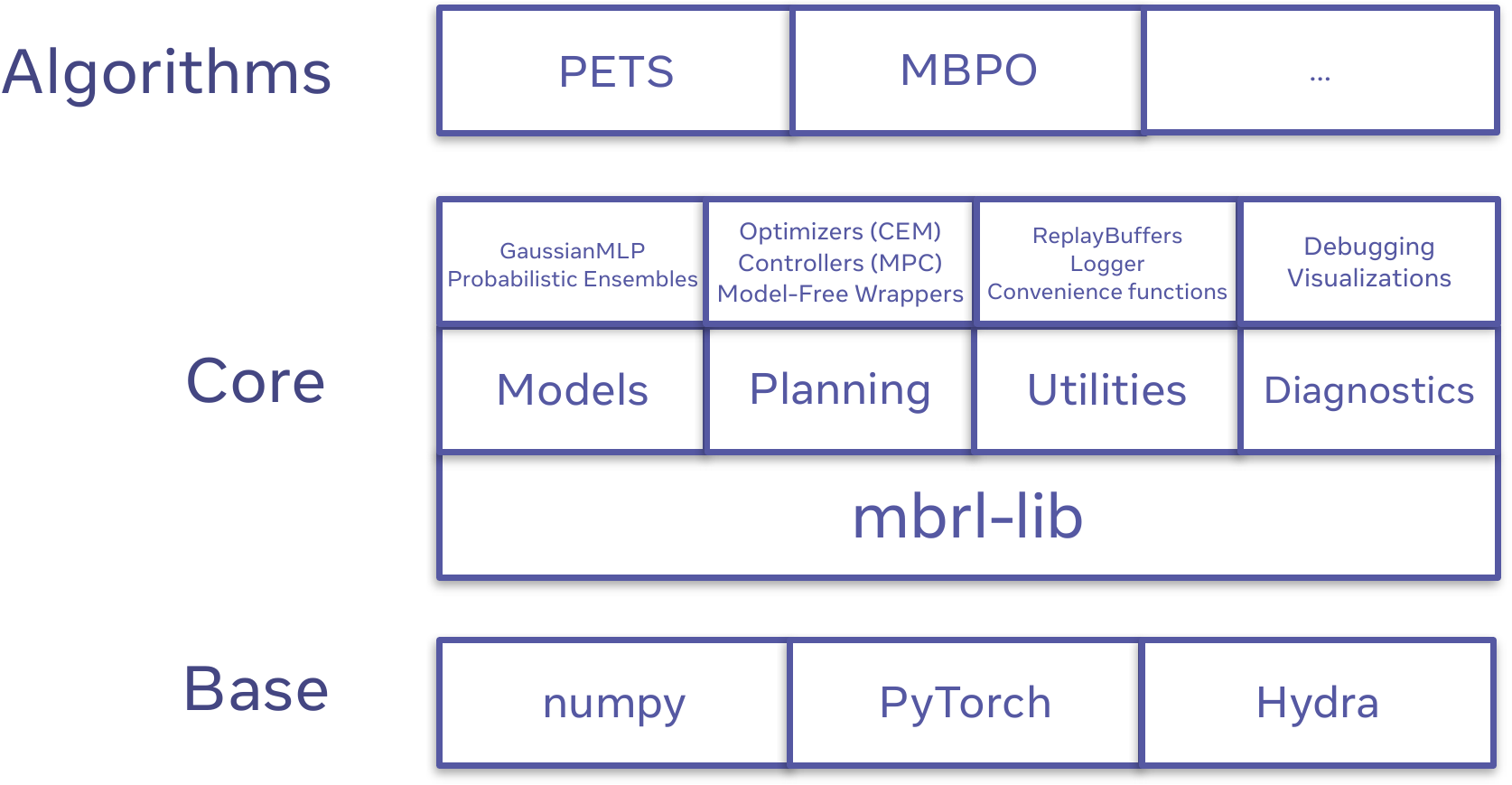}
    \caption{High-level structure of \namelib{}.}
    \label{fig:library_structure}
\end{figure}

Figure~\ref{fig:library_structure} shows the high-level structure of \namelib{}.
The library is built on top of numpy~\citep{van2011numpy} and PyTorch~\citep{pytorch2019} for numeric computation, and Hydra~\citep{Yadan2019Hydra} for managing configurations.
Following the modularity principle, \namelib{} is divided into the following four packages, which are described in more details in the next few subsections:

\begin{itemize}[leftmargin=*,noitemsep,topsep=0pt]
    \item \textbf{mbrl.models:} Dynamics model architectures, trainers, wrappers for data pre- and post-processing.
    \item \textbf{mbrl.planning:} Planners and optimizers of action choices.
    \item \textbf{mbrl.util:} General data-management utilities (\eg, replay buffer, logger) and common operations.
    \item \textbf{mbrl.diagnostics:} Visualizations and tools for diagnosing your models and algorithms.
\end{itemize}

\subsection{Models Package}
The core data abstraction for training models in \namelib{} is \lst{TransitionBatch}, a data class that stores a batch of transitions of observations, actions, next observations, rewards and terminal indicators; the underlying format can be either \lst{numpy} arrays or \lst{torch} tensors.
On top of this data class, the \lst{mbrl.models} package is built around three core classes: \lst{Model}, \lst{ModelTrainer}, and \lst{ModelEnv}.

\subsubsection{Model}
\lst{Model} is the abstract class defining all models that can be trained with \lst{ModelTrainer} and simulated with \lst{ModelEnv}.
We have intentionally kept the required functions and parameters of \lst{Model} interface minimal to account for the variety in choices for modeling dynamics in the literature.
Following the PyTorch convention, all models have a \lst{model.forward(TransitionBatch | torch.Tensor)} method that can return any number of output tensors (\eg, mean, log-variance, log-probabilities).
Additionally, the interface defines two abstract methods that all subclasses must implement:

\begin{itemize}[leftmargin=*,noitemsep,topsep=0pt]
    \item \textbf{\lst{loss(TransitionBatch | torch.Tensor, target=Optional[torch.Tensor])}} \, computes and returns a tensor representing the loss associated to the input.
    Using this, \lst{Model} provides a default \lst{update(ModelInput, target=None)} method that updates the model using back-propagation on the result of \lst{loss}. \lst{ModelTrainer} will rely on \lst{model.update(trasition_batch)} during model training. For models that are trained without back-propagation, such as a Gaussian Processes, users can opt to override the \lst{update} method directly.
    \item \textbf{\lst{eval_score(TransitionBatch | torch.Tensor, target=Optional[torch.Tensor])}} \, computes and returns a tensor representing an evaluation score for the model. This method will be used by \lst{ModelTrainer} to compute a validation score.
\end{itemize}

Note that the input for these two methods can be both a \lst{TransitionBatch} or a \lst{torch.Tensor}.
While, the \lst{ModelTrainer} will only send \lst{TransitionBatch} objects during training (in \lst{numpy} format), we allow the model to receive torch tensors if this is more natural (\eg, when using a feed-forward NN). Also, the \lst{target} tensor for \lst{loss} and \lst{eval_score} is defined as an optional argument, since this won't be needed for models taking a \lst{TransitionBatch} input. To solve the data mismatch with the trainer when models using \lst{torch.Tensor} as inputs, we provide model instances that function as data processing wrappers to convert transition data into appropriate input tensors. One such example is \onedmodel{}, which also provides convenient data processing capabilities, such as setting up delta targets as in Eq.~(\ref{eq:delta_obs}), easily toggle between learning rewards or not, provides an input normalizer, as well as the option to add any custom observation pre-processing function.

Listing~\ref{lst:create_gaussian_mlp} exemplifies a typical usage pattern:

\begin{figure}[thp]
\centering

\begin{minipage}{\textwidth}
\begin{minted}[fontsize=\small,frame=single]{python}
import mbrl.models as models

net = models.GaussianMLP(in_size=5, out_size=4, device="cuda:0")
wrapper = models.OneDimTransitionRewardModel(net, target_is_delta=True)
trainer = models.ModelTrainer(wrapper)
trainer.train(*trainer_args)
model_env = models.ModelEnv(env, wrapper, *model_env_args)

\end{minted}
\captionof{listing}{Example of how to train a neural net to model 1-D transitions and rewards a Gaussian distributions in \namelib{}}
\label{lst:create_gaussian_mlp}
\end{minipage}
\end{figure}

Finally, the following methods are used by \lst{ModelEnv} to simulate trajectories in the environment.
They are optional for subclasses---basic default implementations are provided---, but can be useful for algorithms with specialized simulation pipelines (\eg, latent variable models).
They both receive a \lst{[TransitionBatch | torch.Tensor]} input and return an observation to give to agents.
For example, a model could receive a high-dimensional observation (\eg, pixels), and return a low dimensional vector to use for control. Their differences are:
\begin{itemize}[leftmargin=*,noitemsep,topsep=0pt]
    \item \textbf{\lst{reset}:} \, it should use only observation information and initialize any internal state the model needs over a trajectory.
    \item \textbf{\lst{sample}:} \, it will typically use both observation and action information, and does not need to do any initialization.
\end{itemize}

\subsubsection{ModelTrainer}
The model trainer provides a training loop for supervised learning.
In particular, its \lst{train} method repeatedly updates the model for some number of epochs, and keeps track of the best model found during the training.
The training and (optional) validation datasets are provided as objects of type \lst{TransitionIterator}, which decouples the model trainer from the specifics of the replay buffer data storage; this allows users to train models on customized datasets without altering the data collection process.
One such example is the \lst{BootstrapIterator}, which provides transition batches with an additional model dimension that can be used to train ensembles of bootstrapped models; each model's batch is sampled from its own bootstrapped version of the data in the iterator.

Additionally, a common pattern in MBRL methods using ensemble models is to keep track of the best $n$ models of the ensemble.
This is handled by the trainer (using validation scores) if the \lst{Model} instance has defined a \lst{model.set_elite()} method.

Also, note that given the flexibility provided by the use of \lst{TransitionBatch}, it is possible to use \lst{ModelTrainer} to train other types of models, including, for example, imitation learning agents; the only requirements is to define \lst{update()} properly.

\subsubsection{ModelEnv}
A standard approach to use dynamics models is to rollout (or ``imagine'') trajectories of actions using the model.
To facilitate this, \lst{ModelEnv} is a class that wraps the dynamics model into a reinforcement learning environment with an OpenAI's \lst{gym}-like interface~\citep{brockman2016openai}.
It is then easy to use the model for planning with agents of different type, by following the well known \lst{reset()} and \lst{step()} usage pattern, as Listing~\ref{lst:model_env} illustrates.

\begin{figure}[thp]
\centering

\begin{minipage}{\textwidth}
\begin{minted}[fontsize=\small,frame=single]{python}
import gym
import mbrl.models as models
import numpy as np
from mbrl.env.termination_fns import hopper

# Initialize the model
env = gym.make("Hopper-v2")
net = models.GaussianMLP(in_size=14, out_size=12, device="cuda:0")
wrapper = models.OneDimTransitionRewardModel(
    net, target_is_delta=True, learned_rewards=True)
# Construct the model environment
model_env = models.ModelEnv(
    wrapper, *model_env_args, term_fn=hopper)
# Simulate one step of the environment using the model
obs = env.reset()
model_obs = model_env.reset(obs[np.newaxis, :])
action = env.action_space.sample()
next_obs, reward, done, _ = model_env.step(
    action[np.newaxis, :], sample=True)

\end{minted}
\captionof{listing}{Example of use of \lst{ModelEnv}.}
\label{lst:model_env}
\end{minipage}
\end{figure}

The constructor for \lst{ModelEnv} requires specifying a termination function to compute terminal indicators; the user can also pass an optional reward function, to use if rewards are not learned.
The \lst{reset} method receives a batch of observations to start the episode from, and \lst{step} a batch of actions.
When doing a step, the user can also specify if the model should return a random sample (if possible) or not.

\subsection{Planning Package}
The planning package defines a basic \lst{Agent} interface to represent different kinds of decision-making agents, including both model-based planners and model-free agents.
The only method required by the interface is \lst{act(np.ndarray, **kwargs)}, which receives an observation and optional arguments, and returns an action.
Optionally, the user can define a \lst{plan} method, which receives an observation and returns a sequence of actions; \lst{Agent}'s default implementation of \lst{plan} just calls the \lst{act} method.

Currently, we provide the following agents:

\begin{itemize}[leftmargin=*,noitemsep,topsep=0pt]
    \item \textbf{\lst{RandomAgent}:} returns actions sampled from the environments action space.
    \item \textbf{\lst{TrajectoryOptimzerAgent}:} An agent that uses a black-box optimizer to find the best sequence of actions for a given observation; the optimizer choice can be changed easily via configuration arguments.
    The user must provide a function to evaluate the trajectories.
    We currently provide a function that evaluates trajectories using \lst{ModelEnv}, and returns their total predicted rewards, essentially implementing a Model Predictive Control (MPC)~\citep{camacho2013model} agent.
    We also provide a Cross-Entropy Method~\citep{de2005tutorial} (CEM) optimizer.
    \item \textbf{\lst{SACAgent}:} a wrapper for a popular implementation of Soft Actor-Critic~\citep{pytorch_sac,haarnoja2018soft}.
\end{itemize}

\subsubsection{The Cross-Entropy Method (CEM)}
Here we quickly define CEM, which is a popular algorithm for
non-convex continuous optimization \citep{de2005tutorial}.
In general, CEM solves the optimization problem
$\argmin_x f(x)$ for $x\in\sR^n$.
Given a \emph{sampling distribution} $g_\phi$, the hyper-parameters of CEM are
the number of \emph{candidate points} sampled in each iteration $N$,
the number of \emph{elite candidates} $k$ to use to
fit the new sampling distribution to, and
the number of iterations $T$.
The iterates of CEM are the \emph{parameters} $\phi$ of the
sampling distribution.
CEM starts with an \emph{initial} sampling distribution
$g_{\phi_1}(X)\in\sR^n$,
and in each iteration $t$ generates $N$ samples from the domain
$\left[X_{t,i}\right]_{i=1}^N \sim g_{\phi_t}(\cdot)$,
evaluates the function at those points $v_{t,i}\defeq f_\theta(X_{t,i})$,
and re-fits the sampling distribution to the top-$k$ samples by
solving the maximum-likelihood problem
\begin{equation}
  \phi_{t+1} \defeq
    \argmax_\phi \sum_i \mathbbold{1}{\{v_{t,i} \leq \pi(v_t)_k\}} \log g_\phi(X_{t,i}),
   \label{eq:cem-update}
\end{equation}
where the indicator $\mathbbold{1}\{P\}$ is 1 if $P$ is true and 0 otherwise,
$g_\phi(X)$ is the likelihood of $X$ under the distribution
$g_\theta$,
and $\pi(x)$ sorts $x\in\sR^n$ in ascending order so that
$$\pi(x)_1 \leq \pi(x)_2 \leq \ldots \leq \pi(x)_n.$$
We can then map from the final distribution $g_{\phi_T}$ back
to the domain by taking the mean of it,
\ie $\hat x \defeq \E[g_{\phi_{T+1}}(\cdot)]$,
or by returning the best sample.

In MBRL, we define the objective $f(x)$ to be the control
optimization problem in Eq.~(\ref{eq:control}) over
the actions.
MBRL also often uses multivariate isotropic
Gaussian sampling distributions parameterized by $\phi=\{\mu, \sigma^2\}$.
Thus as discussed in,
\eg, \citet{friedman2001elements},
Eq.~(\ref{eq:cem-update}) has a closed-form solution
given by the sample mean and variance of the top-$k$ samples
as $\mu_{t+1} = (1/k)\sum_{i\in\gI_t} X_{t,i}$
and $\sigma^2_{t+1} = (1/k)\sum_{i\in\gI_t} \left(X_{t,i}-\mu_{t+1}\right)^2$,
where the top-$k$ indexing set is $\gI_t= \{i: v_{t,i} \leq \pi(v_t)_k\}$.

\subsection{Utilities}
Following the ease-of-use principle, \namelib{} contains several utilities that provide functionality that's common to many MBRL methods.
For the utilities we heavily leverage \lst{Hydra}~\citep{Yadan2019Hydra} configurations; an example of a typical configuration file (saved in YAML format) is shown in Listing~\ref{lst:cfg_example}, which we will use as a running example.

\begin{figure}[thp]
\centering
\begin{minipage}{\textwidth}
\begin{minted}[fontsize=\small,frame=single]{yaml}
# dynamics model configuration
dynamics_model:
    model:
      _target_: mbrl.models.GaussianMLP
      device: "cuda:0"
      num_layers: 4
      in_size: "???"  # our utilities will complete this info
      out_size: "???"  # our utilities will complete this info
      ensemble_size: 5
      hid_size: 200
      use_silu: true
      deterministic: false
      propagation_method: "fixed_model"

# algorithm specific options
algorithm:
    initial_exploration_steps: 5000
    learned_rewards: false
    target_is_delta: true
    normalize: True

# experiment specific options
overrides:
    env: "gym___Hopper-v2"
    term_fn: "hopper"
    trial_length: 1000,
    num_trials: 125,
    model_batch_size: 256,
    validation_ratio: 0.05

agent:
    # agent's configuration, omitted for space considerations
optimizer:
    # optimizer's configuration for the agent, if it uses one

\end{minted}
\captionof{listing}{Example \lst{Hydra} configuration for \namelib{}. The string "???" is a special Hydra sequence that indicates this should be completed at runtime. Our utilities take care of handling these inputs according to the environment selected.}
\label{lst:cfg_example}
\end{minipage}
\end{figure}

Listing~\ref{lst:using_util} shows examples of some of the utilities provided by \namelib{}.
Function \lst{make_env} in the \lst{util.mujoco} instantiates the environment, supporting both gym and \lst{dm\_control}~\citep{tassa2020dmcontrol}, as well as custom versions of some of these environments, available in \lst{mbrl.env}.
Function \lst{util.create_one_dim_tr_model} creates a base model (here, \lst{mbrl.models.GaussianMLP}) with the correct input/outputs for this experiment, and creates a \onedmodel{} wrapper for it.
Function \lst{create_replay_buffer} creates a replay buffer of appropriate size (other optional configuration options are available), and function \lst{rollout_agent_trajectories} populates the buffer with steps taken in the environment with a random agent. Note also that creation utilities can be given a path to load data from a directory, making it easy to re-use data from previous runs.
Finally, \lst{train_model_and_save_model_and_data} creates \lst{TransitionIterators} for the trainer, normalizes the data if the model provides a normalizer, runs training, and saves the result to the given save directory.

Overall, our utilities replace a lot of the scaffolding associated with running MBRL experiments, and make it easy for users to concentrate on the high-level details of algorithm and model design.

\begin{figure}[thp]
\centering

\begin{minipage}{\textwidth}
\begin{minted}[fontsize=\small,frame=single]{python}
import mbrl.planning as planning
import mbrl.util.common as mbrl_util
import mbrl.util.mujoco as mujoco_util

# Initialize the model, trainer and the model environment
env, term_fn, reward_fn = mujoco_util.make_env(cfg)
obs_shape = env.observation_space.shape
act_shape = env.action_space.shape
# takes care of model input/output size, and
# adding the 1-D model wrapper
model = mbrl_util.create_one_dim_tr_model(cfg, obs_shape, act_shape, model_dir=None)
model_env = mbrl.models.ModelEnv(env, model, term_fn, reward_fn)
trainer = mbrl.models.DynamicsModelTrainer(model)

# create a replay buffer for the run
# automatically sets the size based on the number of trials to run
# and their length, but user can also override with a provided size
replay_buffer = mbrl_util.create_replay_buffer(
    cfg, obs_shape, act_shape, load_dir=None)

# rollout trajectories in the environment and
# optionally populate a replay buffer
mbrl_util.rollout_agent_trajectories(
    env,
    cfg.algorithm.initial_exploration_steps,
    planning.RandomAgent(env),
    {}, # keyword arguments for agent
    replay_buffer=replay_buffer,
)
# create a trajectory optimizer agent that
# evaluates trajectories in the ModelEnv
agent = planning.create_trajectory_optim_agent_for_model(
    model_env, cfg.algorithm.agent)

# create iterators for current buffer data, update normalization stats,
# run training, log results
mbrl_util.train_model_and_save_model_and_data(
    model, trainer, cfg.overrides, replay_buffer, savedir)

\end{minted}
\captionof{listing}{Example of utilities provided by \namelib{}}
\label{lst:using_util}
\end{minipage}
\end{figure}

\subsection{Diagnostics Package}
\begin{figure}
    \centering
    \includegraphics*[width=0.7\textwidth]{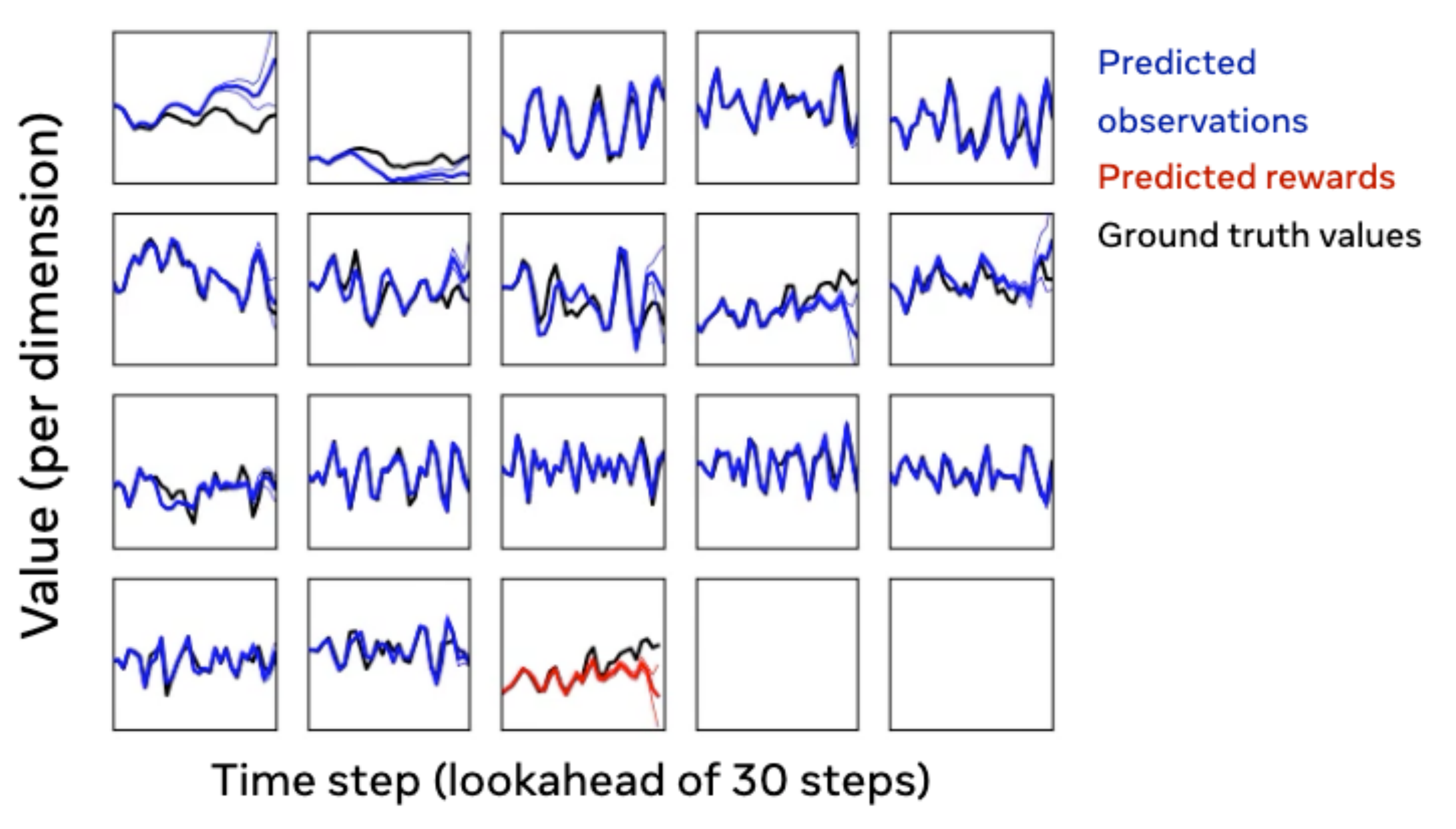}
    \caption{Example result of \lst{mbrl.diagnostics.Visualizer}'s output on an MPC controller with CEM for HalfCheetah. Each subplot corresponds to predictions for a model dimension for 30 time steps, compared with the result of applying the same actions to the true environment. Uncertainty can be visualized via multiple model samples (best visible in the upper left subplot).}
    \label{fig:visualizer}
\end{figure}

\begin{figure}[t!]
    \centering
    \includegraphics*[width=0.6\textwidth]{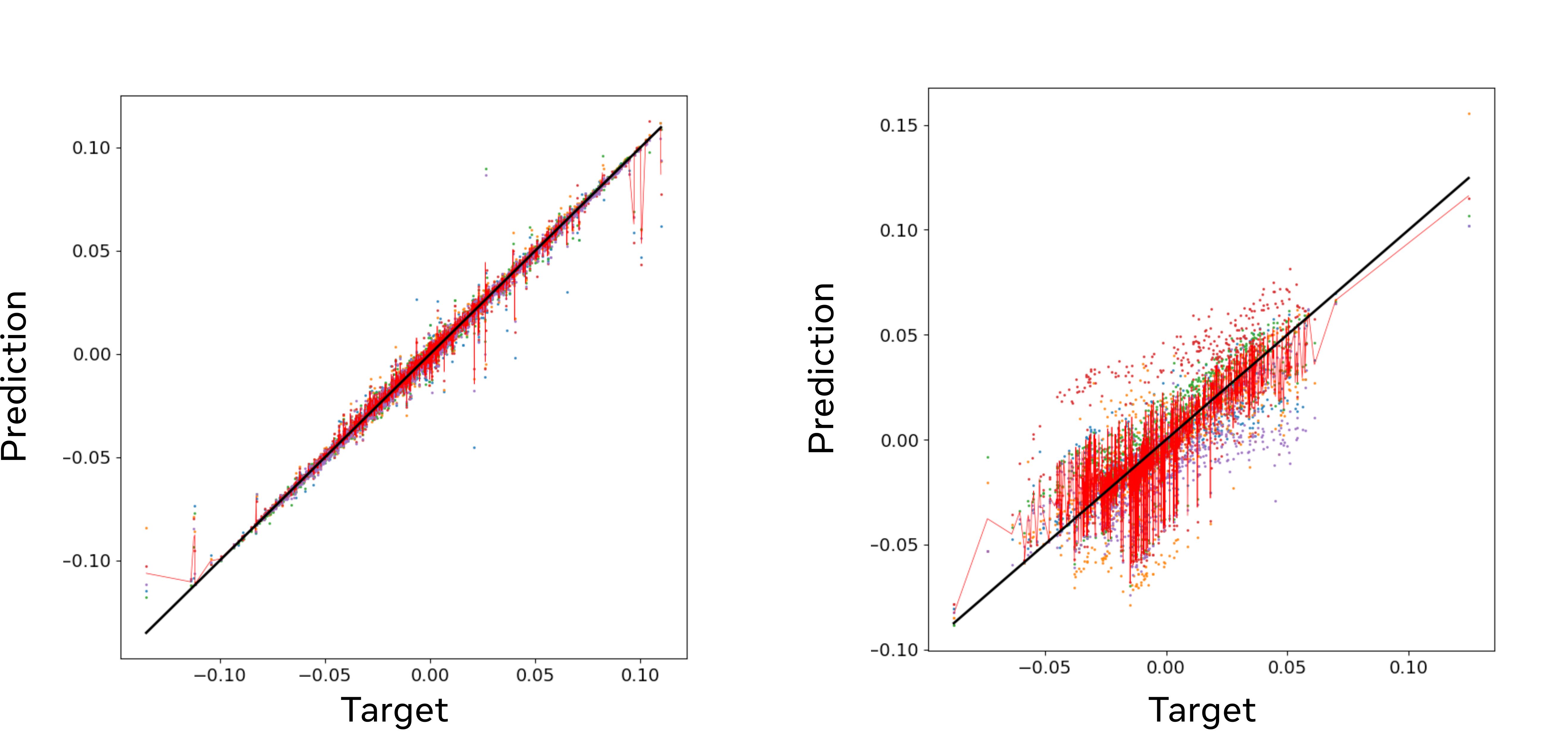}
    \caption{Example result of \lst{mbrl.diagnostics.DatasetEvaluator}'s output. The model was trained on HalfCheetah using MPC control with CEM, the output corresponds to the first observation dimension. Left: model evaluated on the dataset it was trained on. Right: model evaluated on dataset collected with a SAC agent. Note that the predictions of this particular model are bad around optimal trajectories, highlighting problem in learning and data collection.}
    \label{fig:dataset_evaluator}
\end{figure}

The final package is a set of visualization and debugging tools to develop and compare algorithms implemented in \namelib{}.
The current version contains the following diagnostics:
\begin{itemize}[leftmargin=*,noitemsep,topsep=0pt]
    \item \textbf{\lst{Visualizer}} \, Can be used to generate videos of trajectories generated by the agent the model was trained on (and optionally another reference agent). Figure~\ref{fig:visualizer} shows an example output of this tool.
    \item \textbf{\lst{Finetuner}} \, Can be used to train a model (new or saved from a previous run) on data generated by a reference agent. For example, take a model trained with data gathered while doing MPC, and retrain it with data from an optimal controller (\eg, a pre-trained model-free agent). This can be useful to assess if the model is capable of learning correctly around trajectories of interest.
    \item \textbf{\lst{DatasetEvaluator}} \, Takes a trained model and evaluates it in a saved dataset (not necessarily the same one used for training).
    The tool produces plot comparing the predicted values to the targets, one for each predicted dimension.
    An example is shown in  \fig{fig:dataset_evaluator}, where the same model is evaluated in two different datasets.
\end{itemize}

\begin{figure}
    \centering
    \includegraphics[width=0.98\textwidth]{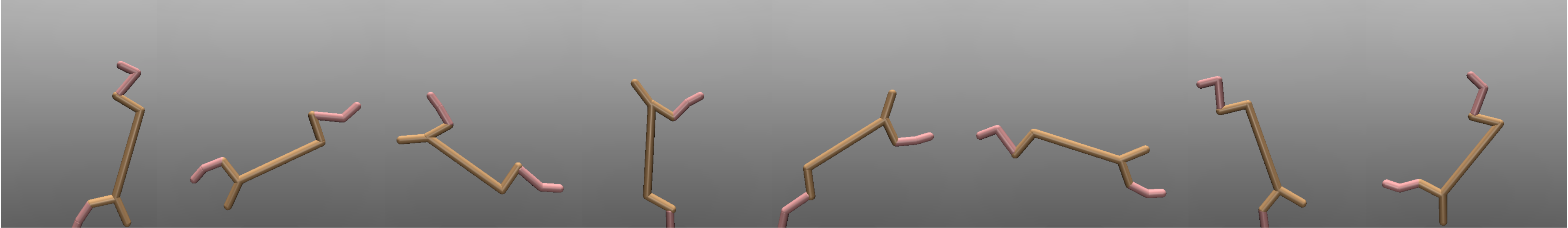}
    \caption{Final frames of an episode run with our implementation of a CEM-based trajectory optimizer on HalfCheetah-v2 environment. The simulation crashed with an overflow error after 679 steps, with an accumulated reward of 20215.78.}
    \label{fig:cheetah_roll}
\end{figure}

We also provide a script for doing CEM control using \lst{TrajectoryOptimizerAgent} on the true environment, which uses Python multiprocessing to launch multiple copies of the environment to evaluate trajectories. This script can be easily extended to other members of \lstinline{mbrl.planning} and can be useful to assess the quality of new optimization algorithms or evaluation functions without any concern over training a perfect model. Figure~\ref{fig:cheetah_roll} illustrates the result of applying this controller to gym's HalfCheetah-v2 environment, which ends up breaking the simulator due to overflow error.

\section{Experimental Results}
\label{sec:result}

	As proof of concept for \namelib{}, we provide implementations for two state-of-the-art MBRL algorithms, namely, PETS~\citep{Chua2018Deep} and MBPO~\citep{janner2019mbpo}. 
We start by briefly describing these algorithms:

\begin{itemize}[leftmargin=*,noitemsep,topsep=0pt]
    \item \textbf{PETS: } The method consists of repeatedly applying two alternating operations: 
    1) train a probabilistic ensemble on all observed transitions, 
    2) run a trial of a trajectory sampling controller (CEM) over model-generated trajectories.  
    \item \textbf{MBPO: } Like PETS, a probabilistic ensemble is trained on all observed transitions. 
    However, rather than using the model for control, short simulated trajectories $\tau$ originating from points in the environment dataset, $\mathcal{D}$, are added to the replay buffer of a SAC agent, which is trained on the simulated data. 
\end{itemize}

Unless otherwise specified, we used ensemble sizes of 7 models trained with Gaussian NLL, as described in \eq{eq:nll}, with predictions using the top 5 ``elite'' models, according to validation score during training. All models predict delta observations as defined in \eq{eq:delta_obs}, and observations are normalized using all the data in the replay buffer before each training episode. Unless otherwise specified, the dynamics model is re-trained on all the available data every 250 steps. For MBPO, we use 20\% of the data for validation, with a new split generated before each training loop starts (as it's done in the original MBPO implementation). For PETS we do not use validation data, following the original implementation, and elites are computed using the training MSE score. 

\begin{figure}
    \centering
    \includegraphics*[width=\textwidth]{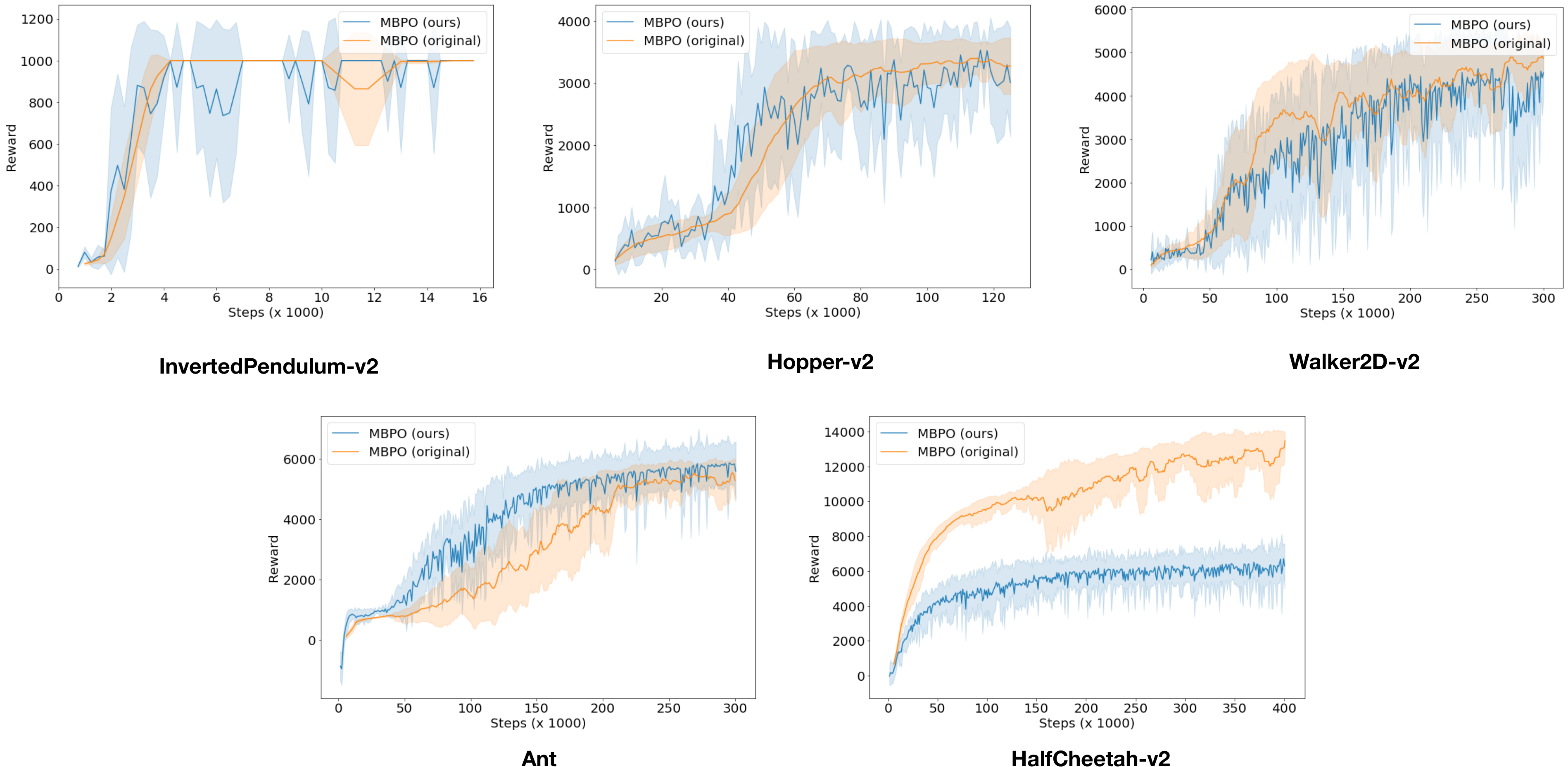}
    \caption{Results of our implementation of MBPO in five Mujoco environments. For the Ant environment, forces are removed from the observations, as was done in the original MBPO implementation.}
    \label{fig:mbpo_results}
\end{figure}

\fig{fig:mbpo_results} shows a comparison of the results obtained with our MBPO implementation and the original on five Mujoco 2.0 environment; the shaded regions correspond to standard deviation using 10 different random seeds (original plots were done with 5 seeds). As the plots show, for the majority of environments our code matches well the original results, and in one case (Ant) the average return observed is higher. A notable exception is HalfCheetah environment, where our results are significantly lower than the original; we are still investigating potential causes for this discrepancy. Our implementation also experiences trouble in the Humanoid-v2 environment, where the SAC losses tend to explode. 

On the other hand, we have had more trouble reproducing the results of the original PETS implementation and have thus omitted a direct comparison with the original, as the large gap in performance is not particularly informative. Moreover, the original PETS code uses handcrafted transformations for the inputs and a given reward function, which makes comparison with other methods confusing, so we opted for a more standard learning setup in our experiments. 

\begin{figure}
    \centering
    \includegraphics*[width=\textwidth]{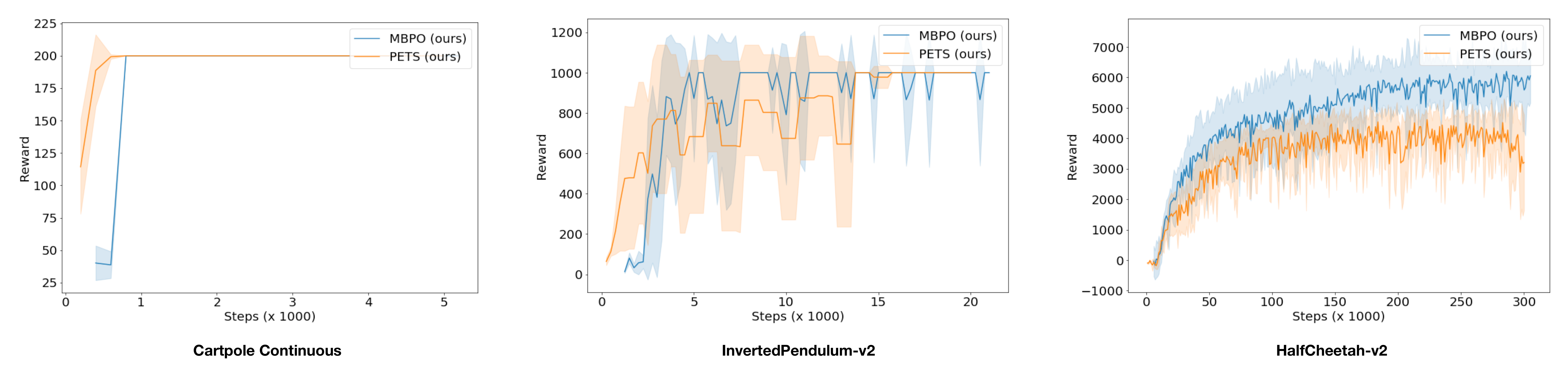}
    \caption{Comparison between the \namelib{} implementations of PETS and MBPO in three different environments.}
    \label{fig:pets_results}
\end{figure}

\fig{fig:pets_results} shows a comparison of our implementations of PETS and MBPO in three environments, namely, Inverted Pendulum, HalfCheetah, and a continuous version of CartPole; the shaded region corresponds to standard deviation over 10 seeds. The performance in Cartpole and InvertedPendulum is close between the two algorithms, although PETS tends to be more unstable in InvertedPendulum. Interestingly, we found that deterministic ensembles allowed PETS to learn faster in CartPole environment; for the other environments, a probabilistic ensemble worked better. 

However, the performance in HalfCheetah is consistently lower in PETS than in MBPO. Note that, in contrast to the original PETS implementation, for this experiments we used unmodified observations for all environments and learned rewards. Adding these modifications improves performance in the HalfCheetah environment, but not enough to match that of the original Tensorflow implementation~\citep{zhang2021importance}.

\section{Conclusion}
\label{sec:conclusion}

	We introduce \namelib{} -- the first PyTorch library dedicated to model-based reinforcement learning.
This library has the double goal of providing researchers with a modular architecture that can be used for quickly implement, debug, evaluate and compare new algorithms, and to provide easy-to-use state-of-the-art implementations for non-expert users.
We open-source \namelib{} at \website{}, with the aim of foster and support the model-based reinforcement learning community.



\begin{ack}
R.C. thanks Omry Yadan for useful discussions regarding software engineering and for work on an early version of MBRL library. L.P. thanks Shagun Sodhani, Olivier Delalleau, Po-Wei Chou, and Raghu Rajan for early feedback on the library, Michael Janner for answering questions about MBPO, and Xingyu Lin, whose PyTorch MBPO implementation was a great resource for debugging our own. 


\end{ack}


\bibliographystyle{abbrvnat}
\bibliography{paper}

\end{document}